\documentclass[11pt]{article}

\usepackage[final]{acl}

\usepackage{times}
\usepackage{latexsym}

\usepackage[T1]{fontenc}
\usepackage[utf8]{inputenc}

\usepackage{microtype}

\usepackage{inconsolata}

\usepackage{graphicx}

\usepackage{amsmath}
\usepackage{amssymb}
\usepackage{amsthm}
\usepackage{booktabs}
\usepackage{multirow}
\usepackage{xcolor}
\usepackage{algorithm}
\usepackage{algpseudocode}
\usepackage[normalem]{ulem}
\usepackage{tikz}
\usetikzlibrary{arrows.meta,shapes.geometric,positioning,calc}

\title{CRISP: Cliff-awaRe Input-adaptive Sparse Prefilling\\
       with Structural-Mass-Motivated Routing}

\author{
  \textbf{Huy Huu Nguyen}\textsuperscript{1} \quad
  \textbf{Chien Van Nguyen}\textsuperscript{1} \quad
  \textbf{Franck Dernoncourt}\textsuperscript{2} \quad
  \textbf{Ryan A. Rossi}\textsuperscript{2} \\
  \textbf{Linh Ngo Van}\textsuperscript{3} \quad
  \textbf{Jieyang Chen}\textsuperscript{1} \quad
  \textbf{Thien Huu Nguyen}\textsuperscript{1} \\[2pt]
  \textsuperscript{1}University of Oregon \quad
  \textsuperscript{2}Adobe Research \quad
  \textsuperscript{3}Hanoi University of Science and Technology \\[2pt]
  \texttt{\{huy, chienn, jieyang, thienn\}@uoregon.edu} \\
  \texttt{\{dernonco, ryrossi\}@adobe.com} \quad
  \texttt{linhnv@soict.hust.edu.vn}
}

\begin{document}
\maketitle

%───────────────────────────────────────────────────────────────────────────────
\begin{abstract}
The attention prefilling phase of long-context LLM inference scales quadratically, making self-attention a severe computational bottleneck.
Traditional sparse attention methods mitigate this through fixed patterns or offline profiling, but lack the flexibility to adapt to input-dependent attention structure.
Recent dynamic methods address this by routing heads to sparse patterns in real-time, but rely on indirect routing proxies with overhead and budget allocation mechanisms that overlook the post-softmax mass hierarchy.
We present \textbf{CRISP} (Cliff-awaRe Input-adaptive Sparse Prefilling), which identifies and addresses two structural challenges in this dynamic routing paradigm.
First, we show that the routing decision can be read directly off the structure of the proxy attention map.
We replace the Jensen-Shannon Divergence (JSD) routing with $C_{\text{struct}}$, a structural proxy that measures mass at Vertical-Slash compatible positions and reproduces JSD's routing decisions while eliminating both the pooled matmul and subsequent KL divergence overhead.
Second, we formalize the post-softmax \textbf{mass cliff} and demonstrate theoretically that strictly cumulative coverage thresholds accumulate $O(n)$ background noise at long contexts.
CRISP navigates this via a sink-aware threshold grounded in the noise floor.
Empirically, across InfiniteBench, RULER and LongBench on two model families, CRISP is the strongest sparse method overall and matches or exceeds exact dense attention on retrieval-heavy benchmarks, recovering up to \textbf{+28.0~pp} on retrieval tasks over baselines and achieving up to a \textbf{5.30$\times$} attention speedup at 512k tokens, driven primarily by our $O(n)$ noise elimination during selection while preserving structural integrity.
\end{abstract}

%───────────────────────────────────────────────────────────────────────────────
% figure

\begin{figure*}[t]
  \centering
  \includegraphics[width=\linewidth]{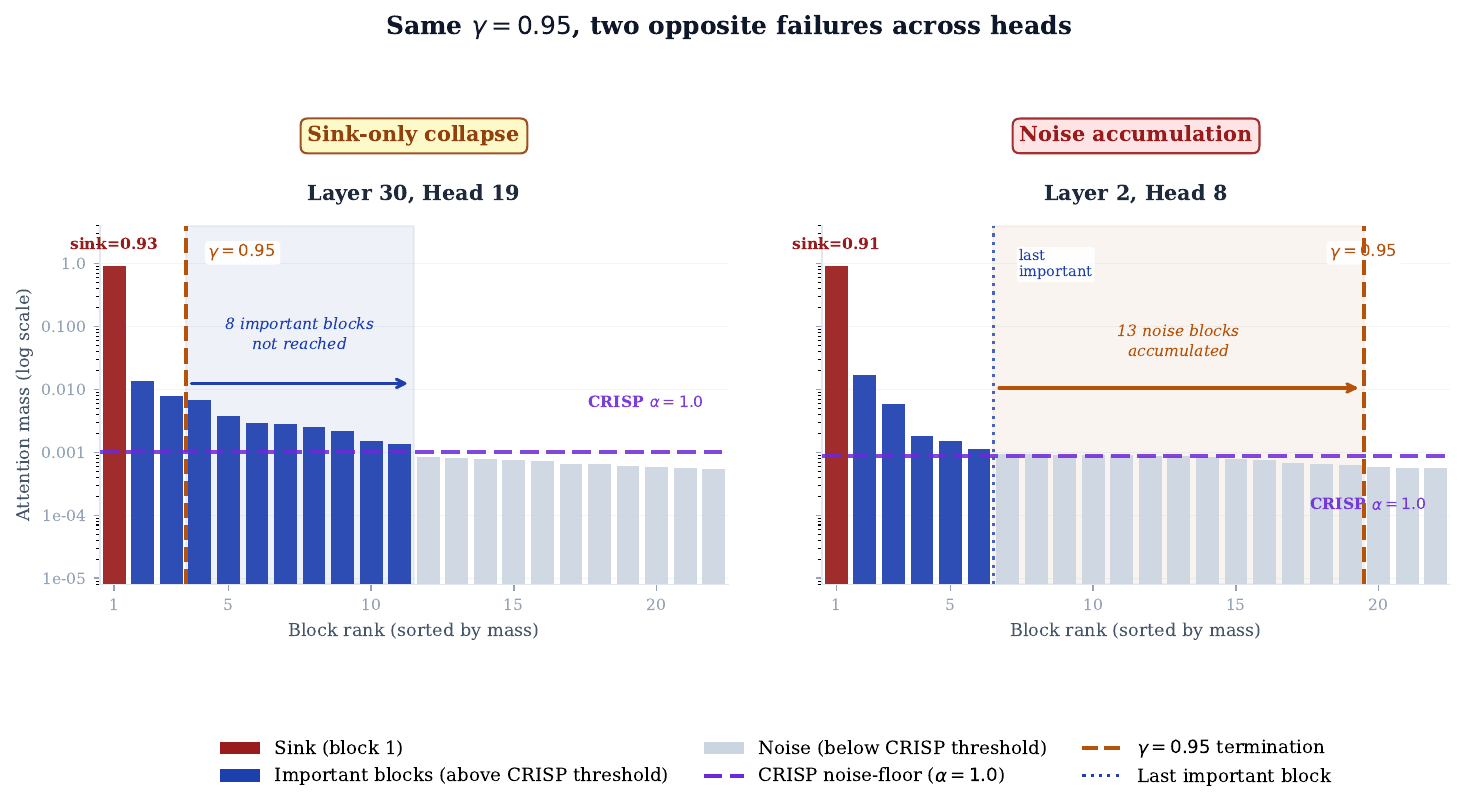}
  \caption{Post-softmax attention mass distribution, illustrating the two opposite failure modes of cumulative thresholding ($\gamma{=}0.95$).
\textbf{Left (sink-only collapse):} high sink mass causes $\gamma{=}0.95$ to terminate inside the sink, leaving important blocks unreached.
\textbf{Right (noise accumulation):} when sink mass is distributed, $\gamma{=}0.95$ exhausts the important band and accumulates 13 noise blocks.
The CRISP noise-floor ($\alpha{=}1.0$, dotted purple) correctly separates signal from noise in both cases.
Additional head examples are provided in Appendix~\ref{app:extra_cliffs}.}
  \label{fig:cliff_histogram}
\end{figure*}

%───────────────────────────────────────────────────────────────────────────────

%───────────────────────────────────────────────────────────────────────────────
\section{Introduction}
\label{sec:intro}

The prefilling phase of long-context LLM inference scales quadratically, making self-attention a severe bottleneck.
Traditional sparse attention uses fixed patterns \citep{child2019generating, zaheer2020bigbird} or offline profiling \citep{jiang2024minference}, lacking the flexibility to adapt to input-dependent attention structures.
To address this, state-of-the-art methods like \textbf{FlexPrefill} \citep{lai2025flexprefill} introduce \emph{dynamic routing}, categorizing heads in real-time to allocate compute budgets.
Specifically, it uses Jensen-Shannon Divergence~(JSD) for pattern routing and a cumulative coverage threshold~$\gamma$ for index selection.
However, analyzing their theoretical foundations reveals principled limitations in both mechanisms.

\paragraph{JSD is an indirect routing signal.}
FlexPrefill sends each head down one of two paths: \emph{Vertical-Slash}~(VS) for heads whose attention is sharply concentrated, \emph{Pooled-Estimation}~(PE) for heads whose attention is spread out.
What the router needs is therefore a cheap read on how concentrated a head is.
JSD obtains one indirectly, by building a second, pooled estimate of the head's attention and measuring how far it diverges from the per-query one, routing to VS above a threshold~$\tau$.
That estimate exists only for the comparison, and costs a matmul and a softmax that nothing else in the method needs.
We take a shortcut instead: in these models a concentrated head places its mass at structurally predictable positions---the attention sinks and the local recency window---so the mass sitting there is itself a read on concentration.
$C_{\text{struct}}$ measures exactly that, costs nothing beyond what index selection already computes, and agrees with JSD's routing decision on $94.0\%$ (Llama) and $88.1\%$ (Qwen) of measured heads.

\paragraph{Cumulative thresholds cannot navigate the mass cliff.}
On VS heads, softmax amplification produces a rigid mass hierarchy: architectural sinks (might absorb up to $>90\%$ mass), task-relevant signal blocks, and near-zero background noise \citep{deng2024attention}.
We term the sharp boundary between signal and noise the \emph{mass cliff}.
Cumulative $\gamma$-thresholding accumulates mass indiscriminately across this boundary, producing two inherent structural failure modes depending on where $\gamma$ falls relative to the sink mass: \emph{sink-only collapse}, in which selection terminates before reaching signal, and \emph{residual noise accumulation}, in which $O(n)$ near-zero background tokens are collected to satisfy the residual threshold.
Tuning $\gamma$ cannot resolve this: increasing coverage merely pushes selection deeper into noise.
We replace $\gamma$ on VS heads with a sink-aware threshold grounded in the noise floor, which explicitly separates signal from architectural noise without empirical calibration.

\paragraph{Contributions.} (1) We give a \textbf{structural account} of VS/PE routing and show empirically that FlexPrefill's JSD signal tracks the same head-level distinction (\S\ref{sec:routing_logic}).
(2) We introduce $C_{\text{struct}}$, a \textbf{structural-mass proxy} that removes the matmul and KL divergence required by JSD-based routing while reproducing its decisions (\S\ref{sec:cstruct}).
(3) We identify the \textbf{post-softmax mass cliff} and provide an asymptotic analysis showing that coverage-based thresholds inherently accumulate $O(n)$ noise at scale (\S\ref{sec:failure2}).
(4) We introduce \textbf{CRISP}, a method using sink-aware thresholding that achieves parity with dense attention and provides up to $5.30\times$ speedups, driven entirely by resolving the $O(n)$ noise accumulation bottleneck (\S\ref{sec:experiments}).

%───────────────────────────────────────────────────────────────────────────────
% figure

% ---- Ultra-Compact Pipeline overview figure ----
\begin{figure*}[t]
\centering
\begin{tikzpicture}[
  % Tightened spacing for maximum horizontal efficiency
  node distance=0.35cm and 0.45cm,
  box/.style={rectangle, rounded corners=2pt, draw, thick,
              minimum height=0.7cm, align=center, font=\footnotesize},
  decision/.style={diamond, draw, thick, aspect=2.0,
                   minimum height=0.7cm, align=center, font=\footnotesize},
  arrow/.style={-{Stealth[length=4pt]}, thick},
  label/.style={font=\scriptsize\itshape},
]

% -- nodes (shrunk widths and simplified text) --
\node[box, fill=gray!12, minimum width=1.3cm] (input)
  {$Q,K,V$};
\node[box, fill=blue!10, minimum width=2.5cm, right=of input] (proxy)
  {Proxy $\hat{A}$\\[-1pt]
   \scriptsize$\hat{A}\!\leftarrow\!\text{Softmax}(\frac{\hat{Q}K^\top}{\sqrt{d}})$};
\node[box, fill=blue!10, minimum width=2.4cm, right=of proxy] (cstruct)
  {Compute $C_{\text{struct}}$\\[-1pt]
   \scriptsize mass at sink/local};
\node[decision, fill=yellow!20, minimum width=2.0cm, right=of cstruct] (gate)
  {$C_{\text{str}} \geq \tau_{\text{px}}$?};

% VS branch
\node[box, fill=orange!18, minimum width=2.4cm,
      above right=0.2cm and 0.5cm of gate] (vs)
  {\textbf{VS path}\\[-1pt]
   \scriptsize Sink-aware $\alpha$-thr.};

% PE branch
\node[box, fill=green!15, minimum width=2.4cm,
      below right=0.2cm and 0.5cm of gate] (pe)
  {\textbf{PE path}\\[-1pt]
   \scriptsize $\gamma$-cumsum};

% Output (merged y-offset)
\node[box, fill=gray!12, minimum width=2.0cm,
      right=0.5cm of vs, yshift=-0.75cm] (out)
  {Sparse Out\\[-1pt]
   \scriptsize $A(Q,K,V,\mathcal{S})$};

% -- arrows --
\draw[arrow] (input)   -- (proxy);
\draw[arrow] (proxy)   -- (cstruct);
\draw[arrow] (cstruct) -- (gate);

\draw[arrow] (gate.north) |- (vs.west)
  node[label, pos=0.25, left, xshift=-1pt] {Y};
\draw[arrow] (gate.south) |- (pe.west)
  node[label, pos=0.25, left, xshift=-1pt] {N};

\draw[arrow] (vs.east) -| (out.north);
\draw[arrow] (pe.east) -| (out.south);

% -- Alg labels (moved closer) --
\node[label, below=0.1cm of proxy,    text=blue!60!black]  {Alg 2};
\node[label, below=0.1cm of cstruct, text=blue!60!black]  {Alg 2};
\node[label, above=0.02cm of vs,      text=orange!70!black]{Alg 3};
\node[label, below=0.02cm of pe,      text=green!50!black] {Alg 4};

\end{tikzpicture}
\caption{CRISP routing: Heads compute proxy attention (Alg 2) and route via $C_{\text{struct}}$ to either a VS path using sink-aware thresholding (Alg 3) or a PE path (Alg 4).}
\label{fig:pipeline}
\end{figure*}
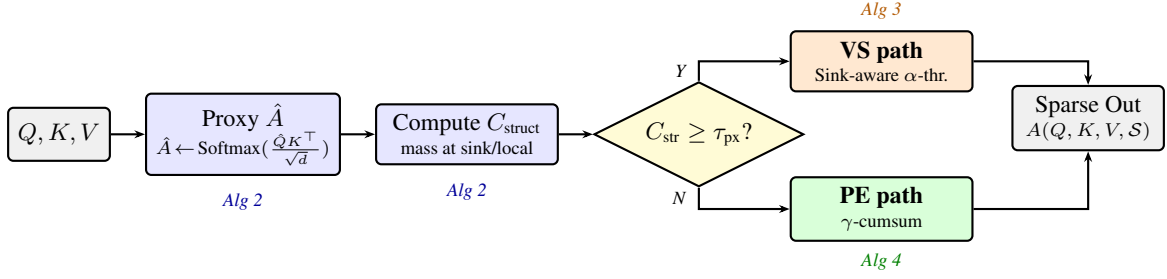

% ---- 4 algorithms in 2x2 layout, mirroring FlexPrefill ----
\begin{figure*}[t]
\begin{minipage}[t]{0.495\linewidth}
\begin{algorithm}[H]
\caption{CRISP Sparse Attention}
\label{alg:main}
\begin{algorithmic}[1]
\State \textbf{Input:} $Q,K,V\in\mathbb{R}^{n\times d}$, $\tau_{\text{proxy}}$, $\gamma$, $\alpha$
\State \textit{\# Route each head}
\State $(\mathit{pattern}, \hat{A}) \leftarrow \textsc{PatSearch}(Q,K,\tau_{\text{proxy}})$
\State \textit{\# Select sparse index set $\mathcal{S}$}
\If{$\mathit{pattern} = \textsc{vs}$}
  \State $\mathcal{S} \leftarrow \textsc{VSIndex}(\hat{A}, \alpha)$
\ElsIf{$\mathit{pattern} = \textsc{pe}$}
  \State $\mathcal{S} \leftarrow \textsc{PEIndex}(Q, K, \gamma)$
\EndIf
\State \textit{\# Sparse attention output}
\State \Return $A(Q,K,V,\mathcal{S})$
\end{algorithmic}
\end{algorithm}
\end{minipage}
\hfill
\begin{minipage}[t]{0.495\linewidth}
\begin{algorithm}[H]
\caption{PatSearch: Pattern Search ($C_{\text{struct}}$)}
\label{alg:pattern}
\begin{algorithmic}[1]
\State \textbf{Input:} $Q, K, \tau_{\text{proxy}}$
\State \textit{\# Representative query subset}
\State $\hat{Q} \leftarrow Q[-b\,{:}\,]$
\State \textit{\# Compute representative attention}
\State $\hat{A} \leftarrow \mathrm{Softmax}(\hat{Q}K^\top/\sqrt{d})$
\State \textit{\# C\textsubscript{struct}: mass at VS-compatible positions}
\State $C_{\text{struct}} \leftarrow \frac{1}{b}\sum_{i}\sum_{j\in\mathcal{S}_{\text{vs}}} \hat{A}[i,j]$
\State \textit{\# Route: constant-time reduction beyond $\hat{A}$}
\If{$C_{\text{struct}} \geq \tau_{\text{proxy}}$}
  \State $\mathit{pattern} \leftarrow \textsc{vs}$
\Else
  \State $\mathit{pattern} \leftarrow \textsc{pe}$
\EndIf
\State \Return $(\mathit{pattern},\, \hat{A})$
\end{algorithmic}
\end{algorithm}
\end{minipage}

\vspace{0.5em}

\begin{minipage}[t]{0.495\linewidth}
\begin{algorithm}[H]
\caption{VS Index (Sink-Aware)}
\label{alg:vs}
\begin{algorithmic}[1]
\State \textbf{Input:} $\hat{A}$, $\alpha$, $k_{\min}$
\State \textit{\# Directional scores, as in FlexPrefill}
\State $a_v \leftarrow \mathrm{colmean}(\hat{A})$;\quad $a_s \leftarrow \mathrm{diagmean}(\hat{A})$
\For{$d \in \{v, s\}$}
  \State $p_d \leftarrow \mathrm{blockpool}(a_d)$ \Comment{reuses $\hat{A}$}
  \State \textit{\# Noise floor over non-retained blocks}
  \State $r_d \leftarrow \max(1 - p_d[0] - p_d[N_b{-}1],\, 0)$
  \State $\mu_d \leftarrow r_d \,/\, (N_b{-}2)$
  \State \textit{\# Budget from the floor, not from coverage}
  \State $k_d \leftarrow |\{j : p_d[j] > \alpha\mu_d\}|$
  \State $k_d \leftarrow \mathrm{clip}(k_d,\, k_{\min},\, N_b)$
  \State $\mathcal{S}_d \leftarrow \text{top-}k_d(p_d)$
\EndFor
\State \Return $\mathcal{S}_v \cup \mathcal{S}_s \cup \{0,\, N_b{-}1\}$
\end{algorithmic}
\end{algorithm}
\end{minipage}
\hfill
\begin{minipage}[t]{0.495\linewidth}
\begin{algorithm}[H]
\caption{PE Index (GlobalAdaptive)}
\label{alg:pe}
\begin{algorithmic}[1]
\State \textbf{Input:} $Q, K, \gamma$
\State \textit{\# Pool over \emph{all} queries (not just $\hat{Q}$)}
\State $\bar{Q} \leftarrow \mathrm{pool}(Q)$;\quad $\bar{K} \leftarrow \mathrm{pool}(K)$
\State $\bar{A} \leftarrow \mathrm{Softmax}(\bar{Q}\bar{K}^\top/\sqrt{d})$
\State \textit{\# Flatten and normalise}
\State $\bar{A} \leftarrow \mathrm{flatten}(\bar{A}\,/\!\sum_{i,j}\bar{A}[i,j])$
\State \textit{\# Sort and $\gamma$-cumsum}
\State $I \leftarrow \mathrm{argsort}(\bar{A})$
\State $K^* \leftarrow \min\!\left\{k : \sum_{i \in I[1:k]} \bar{A}[i] \geq \gamma\right\}$
\State $\mathcal{S} \leftarrow I[1:K^*]$
\State \Return $\mathcal{S}$
\end{algorithmic}
\end{algorithm}
\end{minipage}
\end{figure*}

\begin{figure*}[t]
  \centering
  \includegraphics[width=\linewidth]{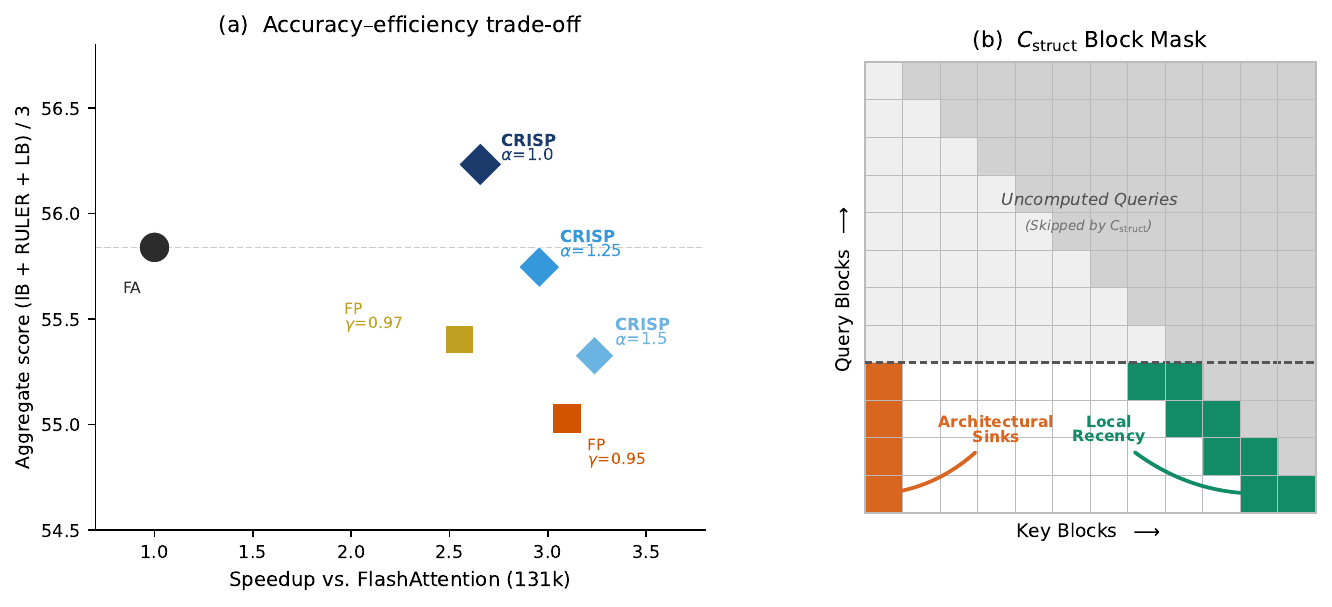} 
  \caption{
  \textbf{(a)} Accuracy-efficiency Pareto plot at 131k tokens.
Each marker shows the aggregate score (mean of InfiniteBench, RULER, and LongBench, averaged over Llama-3.1-8B and Qwen2.5-7B) vs.\ speedup over 
  FlashAttention (FA).
All three CRISP configurations ($\alpha{\in}\{1.0, 1.25, 1.5\}$) Pareto-dominate both FlexPrefill baselines, forming a smooth accuracy-speed frontier.
Notably, pushing FlexPrefill (FP) to higher coverage 
  ($\gamma{=}0.97$) yields minimal accuracy gain but incurs a severe 
  latency penalty, confirming the mass cliff limitation.
\textbf{(b)} The $C_{\text{struct}}$ Block Mask. The proxy attention mass is only 
  measured at structurally stable positions: the Architectural Sinks (orange) and 
  the Local Recency band (green).
Uncomputed queries are skipped for efficiency.
}
  \label{fig:results_overview}
\end{figure*}

\begin{figure*}[t]
  \centering
  \includegraphics[width=\linewidth]{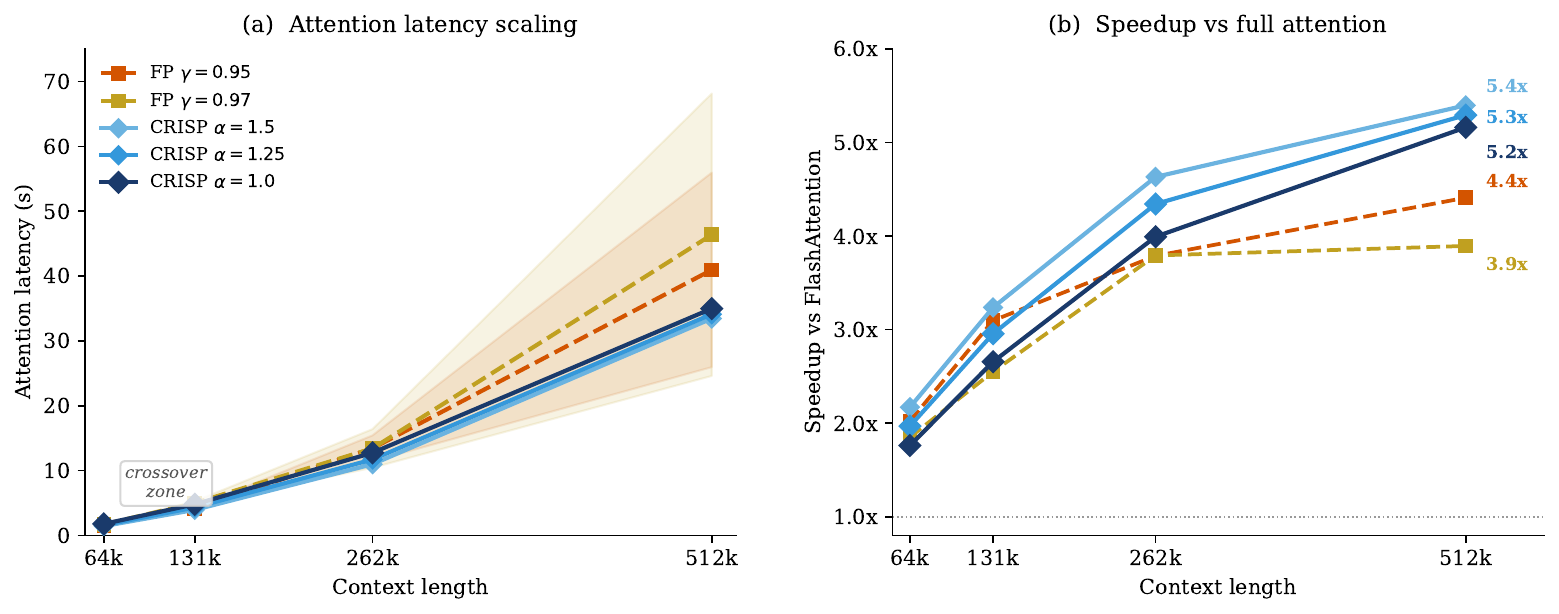}
  \caption{Attention-only latency scaling (64k--512k tokens, single H100,
    mean $\pm$ std shading on FP baselines only for clarity).
\textbf{Left:} absolute latency; FP variance explodes at long contexts
    due to input-dependent cliff position, while all three CRISP configurations
    scale tightly.
\textbf{Right:} speedup over FlashAttention; all CRISP configurations
    overtake their FlexPrefill counterparts beyond 131k,
    reaching $5.17\times$, $5.30\times$, and $5.40\times$ at 512k
    for $\alpha{\in}\{1.0, 1.25, 1.5\}$ respectively.}
  \label{fig:latency}
\end{figure*}

%───────────────────────────────────────────────────────────────────────────────
\section{Background}
\label{sec:background}

\subsection{Sparse Attention}
\label{sec:sparse_attn}
Given sequence length $n$ and hidden dimension $d$, sparse attention restricts computation to an index set $\mathcal{S}$:
\begin{equation}
    A(Q,K,V,\mathcal{S}) = \mathrm{Softmax}\!\left(\tfrac{QK^\top}{\sqrt{d}} + M_\mathcal{S}\right)V
\end{equation}
where $M_\mathcal{S}$ is a mask matrix ($M_{ij}=0$ if $j \in \mathcal{S}_i$, else $-\infty$).
FlexPrefill, the SOTA, formulates $\mathcal{S}$ as the solution to a dual-optimization problem: minimizing $|\mathcal{S}|$ subject to a cumulative mass constraint $\sum_{j \in \mathcal{S}} a_j \geq \gamma$.
This objective implicitly assumes all attention mass contributes equally to output quality, an assumption we challenge in \S\ref{sec:failure2}.

\subsection{FlexPrefill}
\label{sec:flexprefill}
\citet{lai2025flexprefill} routes heads by evaluating a representative query subset $\hat{Q}$.
It computes proxy attention $\hat{A}$ and derives two distributions: a pooled estimate $\bar{a}$ (applying softmax \emph{after} pooling both queries and keys) and a per-query mean $\hat{a}$ (averaging \emph{after} softmax).
Heads are routed to a \emph{Vertical-Slash} (VS) path if $\mathrm{JSD}(\bar{a}, \hat{a}) \geq \tau$, or a \emph{Pooled-Estimation} (PE) path otherwise.
For VS heads, $\hat{A}$ is decomposed into vertical scores (column sums) and slash scores (diagonal sums);
each direction is sorted independently and accumulated until mass $\geq \gamma$, with the final index set formed as $S_v \cup S_s$ plus the first and last blocks.
PE heads are flattened, sorted, and accumulated similarly. We provide the full algorithmic details of this baseline in Appendix~\ref{app:flex_algo}.

\subsection{Attention Entropy}
\label{sec:entropy}
For query $i$, $H(a_i) = -\sum_j a_{ij}\log a_{ij}$. Head entropy is 
$H_{\text{head}} = \frac{1}{|\hat{Q}|}\sum_{i \in \hat{Q}} H(a_i)$. 
As reviewed in Appendix~\ref{app:proof_coverage}, head entropy is 
closely related to the minimum number of tokens required to achieve 
coverage $\gamma$, which motivates the routing dichotomy of 
\S\ref{sec:routing_logic}.

% =============================================================================
% SECTION 3: ENTROPY-THEORETIC ROUTING 
% =============================================================================
\section{Structural-Mass Routing}
\label{sec:routing_logic}

The routing decision follows from a simple observation about 
attention entropy. The information-theoretic link between entropy 
and support size \citep{Campbell1966} tells us that low-entropy 
heads need few tokens for high coverage. A consistent finding 
across prior work is that in autoregressive transformers, these 
tokens occupy structurally predictable positions---architectural 
sinks \citep{xiao2024efficient}, vertical columns, and slash 
diagonals \citep{jiang2024minference, lai2025flexprefill}---giving 
rise to the Vertical-Slash pattern that sparse methods exploit. 
Low-entropy heads are therefore well-suited to the \textsc{vs} 
path, which captures these patterns at fine granularity. 
High-entropy heads lack such structure: mass is spread broadly 
with no dominant pattern for \textsc{vs} to exploit, making 
pooled estimation (\textsc{pe}) the more effective strategy.

This account motivates the routing dichotomy itself; it does not
by itself say which computable quantity should decide it.
\S\ref{sec:failure1} examines what FlexPrefill's JSD signal
measures in practice, and \S\ref{sec:cstruct} replaces it with a
structural quantity that is already available.
The structural half of this account---mass at sinks and
recency---is a property of current sink-having architectures, not
of softmax attention in general
(See Appendix~\ref{app:proof_coverage} for the
information-theoretic grounding of the dichotomy).

\subsection{What JSD Measures}
\label{sec:failure1}

FlexPrefill routes on $\mathrm{JSD}(\bar{a}, \hat{a})$, the divergence between the pooled estimate $\bar{a}$ (softmax applied \emph{after} pooling queries and keys) and the per-query mean $\hat{a}$ (averaging \emph{after} softmax); these two constructions do not agree in general, and it is their disagreement that the routing consumes.
Empirically the disagreement is large exactly on the concentrated heads and small on the diffuse ones, which is what makes it usable for routing---but this is a measured regularity of these models, not a consequence of any property of softmax.
The vertical and slash components of a \textsc{vs} head contribute to it very differently, which offers an intuition for why.
Where queries agree on the same keys---the vertical case---pooling before the softmax and averaging after it yield nearly the same distribution, and the divergence is small.
Where queries peak on \emph{different} keys---the slash case---the two orders come apart: pooling first concentrates $\bar{a}$ on whichever key carries the largest cross-query mean, whereas averaging after leaves $\hat{a}$ spread across the keys individual queries prefer, so the estimates place their mass on different supports and the divergence is large even though every query row is itself concentrated.
Because both structures co-occur in real \textsc{vs} heads, it is the slash component that drives $\mathrm{JSD}$ above $\tau$, while diffuse \textsc{pe} heads produce little divergence from either.
We offer this as an observation consistent with our measurements rather than as a derivation.

Whatever $\mathrm{JSD}$ captures, it is not free.
Computing $\bar{a}$ requires pooling $K$, an additional $O(nd/b)$ matmul (where $b$ is the pooling block size), and a second softmax;
the divergence then adds two logarithm passes and two reductions over the $N_b$ block scores, for every head.

\begin{table}[t]
\centering
\small
\resizebox{\columnwidth}{!}{%
\begin{tabular}{lcc}
\toprule
\textbf{Measurement} & \textbf{Llama-3.1-8B} & \textbf{Qwen2.5-7B} \\
\midrule
Routing agreement with JSD       & 94.0\%  & 88.1\%  \\
Top-1 mass at sink/recency       & 91.3\%  & 98.3\%  \\
\bottomrule
\end{tabular}%
}
\caption{Routing analysis over all attention heads of both models, measured on InfiniteBench inputs.
$C_{\text{struct}}$ reproduces the routing decisions $\mathrm{JSD}$ makes, and concentrated mass sits at the anchor positions it inspects.}
\label{tab:routing_analysis}
\end{table}

\subsection{The Structural Proxy: $C_{\text{struct}}$}
\label{sec:cstruct}

We replace JSD with a direct measurement of VS-compatible mass:
\begin{equation}
    C_{\text{struct}} = \frac{1}{|\hat{Q}|}\sum_{i \in \hat{Q}} \sum_{j \in \mathcal{S}_{\text{anchor}}} \hat{A}[i,j]
\end{equation}
where $\mathcal{S}_{\text{anchor}}$ contains the architectural sinks (first $w_{\text{sink}}{=}128$ tokens) and the local recency window ($w_{\text{local}}{=}128$).
CRISP routes a head to \textsc{vs} if $C_{\text{struct}} \geq \tau_{\text{proxy}}$, and to \textsc{pe} otherwise.

We stress what this quantity is: $C_{\text{struct}}$ measures structural mass at fixed anchor positions, not entropy;
the two coincide empirically because low-entropy heads in current sink-having architectures place their mass at exactly those positions.
Table~\ref{tab:routing_analysis} quantifies this: the top-1 mass block is a sink or recency block for $91.3\%$ (Llama) and $98.3\%$ (Qwen) of measured heads, and the resulting routing agrees with FlexPrefill's JSD decision on $94.0\%$ and $88.1\%$ of them.
That concentration rate is what makes the low-$C_{\text{struct}}$ branch trustworthy: when anchor mass is small, mass is not concentrated elsewhere either.
It is a constant-time indexed slice reduction over $\hat{A}$ already computed for index selection.
Because it strictly evaluates the first $w_{\text{sink}}$ and last $w_{\text{local}}$ tokens per query block, the operation executes in $O(w_{\text{sink}} + w_{\text{local}})$ time, scaling independently of total sequence length $n$, eliminating JSD overhead. Yet, token selection remains constrained by the mass cliff.

% =============================================================================
% SECTION 4: NAVIGATING THE MASS CLIFF
% =============================================================================
\section{Navigating the Mass Cliff}
\label{sec:failure2}

\subsection{The Post-Softmax Token Hierarchy}
On VS heads, softmax creates a three-class hierarchy separated by a hard cliff:
\textbf{Architectural Sinks} often absorb a dominant portion of the total mass;
\textbf{Task-Relevant Signal} holds moderate above-baseline mass; \textbf{Background Noise} carries
near-zero mass that diminishes with sequence length---softmax normalization distributes background mass across $n$ tokens, yielding per-token mass $\epsilon$ that scales as $O(1/n)$ for a fixed pre-softmax gap.
This is consistent with the theoretical framework of \citet{deng2024attention}, which confirms that attention is naturally sparse.

\subsection{The $O(n)$ Scaling Limitations of $\gamma$}

Early work established that attention sparsity must be dynamic and input-dependent rather than static \citep{liu2021transformeraccelerationdynamicsparse}.
To implement this dynamic routing, recent methods mathematically formulate index selection as a constrained dual-optimization problem: minimizing subset size subject to a cumulative coverage constraint, $\sum a_j \geq \gamma$ \citep{lai2025flexprefill}.
However, this specific optimization objective assumes all attention mass is semantically equivalent.
By failing to account for the post-softmax token hierarchy, strictly cumulative formulations overlook the mass cliff, resulting in two inherent structural failure modes at scale:
 
\paragraph{Sink-Only Collapse.}
When $a_{\text{sink}} \geq \gamma$, selection terminates before any
task-relevant tokens are evaluated.
The error bound is formally satisfied, but
the selected set contains zero semantic information.

\paragraph{Residual Noise Accumulation.}
When $a_{\text{sink}} + a_{\text{important}} < \gamma$, the algorithm crosses
the cliff into background noise, collecting $O(\gamma_{\text{residual}}/\epsilon)
\propto O(n)$ noise tokens (see Appendix~\ref{app:cliff}).

\paragraph{$\gamma$-tuning cannot escape the cliff.}
Figure~\ref{fig:cliff_histogram} illustrates both failure modes 
empirically. These are parametrically irreparable: increasing 
$\gamma$ crosses the cliff deeper into noise. 
Table~\ref{tab:summary} confirms this---$\gamma{=}0.97$ 
degrades some benchmarks while marginally improving others, with 
inconsistent effects across models and tasks. This is precisely 
what the mass cliff predicts: the effect depends on where sink 
mass falls relative to the threshold, not on coverage quality. 
CRISP avoids this dependency by grounding selection in the noise 
floor rather than a coverage target.

\subsection{Fix: Sink-Aware Thresholding}
\label{sec:sink_aware_fix}
Throughout, the proxy attention map is partitioned into $N_b = \lceil n / B \rceil$ blocks of $B{=}128$ tokens, so the anchor windows of \S\ref{sec:cstruct} ($w_{\text{sink}} = w_{\text{local}} = 128$) are exactly the first and last block.
We exclude these always-retained blocks---the first block (architectural sinks)
and the last block (local recency, always retained following FlexPrefill)---and
establish a baseline expected mass over the remaining $N_b - 2$ blocks:
\begin{equation}
    \mu = \frac{\max(1 - a_{\text{first}} - a_{\text{last}},\; 0)}{N_b - 2}
\end{equation}
A block is selected if and only if it exceeds the expected background mass:
\begin{equation}
    \text{block } j \text{ is important} \iff p_j > \alpha \cdot \mu
\end{equation}
At $\alpha{=}1.0$, the threshold equals the expected background 
mass---any block above the mean carries above-average signal, 
making $\alpha{=}1.0$ a calibration-free default. The budget $k$ 
emerges from task complexity: focused tasks select fewer blocks, 
dense tasks select more. Unlike $\gamma$, the threshold correctly
lowers $\mu$ as sinks grow, maintaining sensitivity to signal regardless of
sink mass.

Note that for high-entropy PE heads, mass is diffuse and lacks a sharp structural hierarchy.
Consequently, PE heads do not suffer from the same severe mass cliff as VS heads, allowing CRISP to safely retain $\gamma$-cumsum approximation for the PE path without accumulating disproportionate noise (Algorithm~\ref{alg:pe});
Appendix~\ref{app:pe_heads} shows the corresponding mass profiles.

What does \emph{not} change is the decomposition itself.
CRISP still computes both directions: the vertical score $a_v$ (column means of $\hat{A}$) and the slash score $a_s$ (diagonal means) are still computed separately, thresholded separately, and unioned into a single block index set (Algorithm~\ref{alg:vs}; cf.\ Algorithm~\ref{alg:flex_vs} in Appendix~\ref{app:flex_algo}).
What changes is the \emph{budget rule} applied within each direction: the cumulative criterion $\sum a_j \geq \gamma$ is replaced by the noise floor $\alpha\mu_d$, so the number of selected blocks follows from how many blocks carry above-background mass rather than from a coverage target.

%───────────────────────────────────────────────────────────────────────────────
\section{Experiments}
\label{sec:experiments}

\paragraph{Setup.}
We evaluate on Meta-Llama-3.1-8B-Instruct \citep{grattafiori2024llama3herdmodels} and Qwen2.5-7B-Instruct \citep{qwen_qwen25_2025} across
InfiniteBench~\citep{zhang-etal-2024-bench} (131K), RULER~\citep{hsieh2024ruler}
(4K--131K), and LongBench~\citep{bai-etal-2024-longbench} (4K--16K).
For InfiniteBench, inputs are truncated to 131K tokens to align with the models' maximum supported context windows.
CRISP uses a universal configuration:
$\tau_{\text{proxy}}{=}0.2$, $\gamma{=}0.95$ (PE only), $\alpha{\in}\{1.0,
1.25\}$ for the VS path, block size $B{=}128$, and a minimum budget of
$k_{\min}{=}1024$ tokens (8 blocks) per direction, inherited from FlexPrefill.
We report $\alpha{=}1.0$ as the primary recommended
configuration (highest accuracy) and $\alpha{=}1.25$ as the speed-focused
configuration.
To verify that the theoretically motivated $\alpha{=}1.0$ is indeed the correct operating point, we additionally evaluate $\alpha{=}1.5$ (beyond the noise floor) in the sensitivity analysis (Appendix~\ref{app:hyperparams}), which confirms monotonic degradation and demonstrates graceful performance even past the principled default.

Baselines are: \textbf{FlashAttention}~\citep{dao2024flashattention}
(full attention upper bound, implemented as standard exact dense attention); \textbf{MInference}~\citep{jiang2024minference}
(offline sparse patterns, strongest prior method);
and \textbf{FlexPrefill} at
its recommended $\gamma{=}0.95$~\citep{lai2025flexprefill} and additionally at
$\gamma{=}0.97$, an exploratory higher-coverage setting from their ablation,
included as the empirical test of the mass cliff prediction.
Latency is measured as attention-only time on a single NVIDIA H100 80GB,
isolating the operation that CRISP modifies.

\begin{table}[h]
\centering
\small
\resizebox{\columnwidth}{!}{%
\begin{tabular}{lccc}
\toprule
\textbf{Benchmark} & \textbf{Method} &
    \textbf{Llama-3.1-8B} & \textbf{Qwen2.5-7B} \\
\midrule
\multirow{6}{*}{InfiniteBench (131K)}
    & FlashAttn (FA)     & 48.6 & 24.0 \\
    & MInference         & 36.5          & 25.1 \\
    & FP $\gamma{=}0.95$ & 47.4          & 25.5 \\
    & FP $\gamma{=}0.97$ & 47.3\textcolor{red}{$\downarrow$} & 26.7 \\
    & CRISP $\alpha{=}1.25$ & \underline{48.0} & \underline{28.4} \\
    & CRISP $\alpha{=}1.0$  & \textbf{48.7}\textcolor{blue}{$\uparrow$FA} & \textbf{28.7}\textcolor{blue}{$\uparrow$FA} \\
\midrule
\multirow{6}{*}{RULER (4K--131K)}
    & FlashAttn (FA)     & 89.02          & 75.84 \\
    & MInference         & 86.14          & 72.44 \\
    & FP $\gamma{=}0.95$ & \textbf{89.22} & 75.52 \\
    & FP $\gamma{=}0.97$ & \underline{89.14}\textcolor{red}{$\downarrow$} & \underline{75.80} \\
    & CRISP $\alpha{=}1.25$ & 88.62        & \underline{75.80} \\
    & CRISP $\alpha{=}1.0$  & 88.82        & \textbf{76.20}\textcolor{blue}{$\uparrow$FA} \\
\midrule
\multirow{6}{*}{LongBench (4K--16K)}
    & FlashAttn (FA)     & 48.82 & 48.80 \\
    & MInference         & 41.40          & 38.95 \\
    & FP $\gamma{=}0.95$ & 46.92 & 45.57 \\
    & FP $\gamma{=}0.97$ & \underline{47.12} & 46.45 \\
    & CRISP $\alpha{=}1.25$ & 46.92 & \underline{46.76} \\
    & CRISP $\alpha{=}1.0$  & \textbf{47.77} & \textbf{47.23} \\
\bottomrule
\end{tabular}%
}
\caption{Summary results. FP = FlexPrefill; FA = FlashAttention (full attention).
The red arrow (\textcolor{red}{$\downarrow$}) denotes performance degradation when increasing FP coverage from 0.95 to 0.97.
CRISP $\alpha{=}1.0$ achieves parity with or exceeds exact dense attention (\textcolor{blue}{$\uparrow$FA}) on retrieval-heavy tasks and is the top-performing sparse method overall.}
\label{tab:summary}
\end{table}

\subsection{Retrieval Accuracy Recovery}
\label{sec:retrieval}

Table~\ref{tab:retrieval} shows double-digit accuracy gains on retrieval tasks.
These are empirical proof of sink-only collapse: $\gamma$ terminates selection
on sink mass before evaluating task-relevant tokens.
These gains are
structurally unrecoverable by tuning $\gamma$---as confirmed by
Table~\ref{tab:summary}, which shows $\gamma{=}0.97$ provides no
retrieval improvement over $\gamma{=}0.95$ while costing additional latency.
The Qwen2.5 passkey gain (+28.0pp) indicates particularly heavy sink
concentration in that architecture.

\begin{table}[h]
\centering
\small
\resizebox{\columnwidth}{!}{%
\begin{tabular}{llcc}
\toprule
\textbf{Task} & \textbf{Benchmark} &
    \textbf{Llama $\Delta$} & \textbf{Qwen $\Delta$} \\
\midrule
KV Retrieval           & InfiniteBench & +17.8pp  & --\textsuperscript{$*$} \\
Passkey                & InfiniteBench & 0.0pp    & +28.0pp \\
Passage Retrieval (En) & LongBench     & +12.50pp & +13.50pp \\
Passage Retrieval (Zh) & LongBench     & +3.16pp  & +4.50pp \\
RULER 65K              & RULER 
        & +2.65pp  & +1.45pp \\
\bottomrule
\end{tabular}%
}
\caption{Retrieval accuracy recovery vs FlexPrefill $\gamma{=}0.95$.
\textsuperscript{$*$}Qwen2.5 baseline is near-zero on this task; difference
is not meaningful.}
\label{tab:retrieval}
\end{table}

\subsection{General Performance}
\label{sec:general}

Table~\ref{tab:infinitebench} shows full InfiniteBench results.
CRISP
$\alpha{=}1.0$ achieves parity with exact dense attention on both models (Llama: 48.7
vs 48.6; Qwen: 28.7 vs 24.0), a result consistent with the theoretical
prediction that full attention must aggregate over all tokens including
architectural sink noise, whereas CRISP's sink-aware selection produces a
cleaner signal representation.
The mass cliff that constrains FlexPrefill
applies equally to dense attention; CRISP is the first method to explicitly
navigate it.

MInference scores 36.5 on Llama and 25.1 on Qwen, well below FlexPrefill
(47.4 / 25.5), consistent with offline patterns failing to adapt to
input-dependent sink mass.

LongBench results (Table~\ref{tab:longbench}) show MInference loses 5--8pp
vs FlexPrefill, while CRISP gains +0.85pp and +1.66pp over FP $\gamma{=}0.95$.
Notably, even when FlexPrefill is pushed to a higher-coverage regime ($\gamma{=}0.97$), 
it achieves only 47.12 on Llama---still trailing CRISP's 47.77---while incurring a 
massive latency regression (detailed in \S\ref{sec:latency}).
This confirms that coverage-based methods cannot simply parameter-tune their way to CRISP's accuracy 
level;
the noise ingestion fundamentally caps their structural integrity.

\paragraph{RULER on Llama.}
CRISP shows --0.40pp aggregate on Llama RULER vs $\gamma{=}0.95$, while
MInference loses --3.08pp.
The modest regression is mechanistically expected:
RULER's synthetic aggregation tasks require broad coverage of mid-range
attention blocks, and the sink-aware threshold at $\alpha{=}1.0$ occasionally
treats these borderline blocks as noise when their mass falls near $\mu$.
This is a genuine precision-coverage tradeoff --- the same conservatism that
eliminates noise on retrieval tasks marginally under-selects on aggregation.

On Qwen, where sink mass distribution is less concentrated, CRISP wins RULER
outright (76.20 vs 75.52 vs 72.44), confirming the effect is
model-architecture-dependent rather than a systematic limitation.

\subsection{Latency Scaling}
\label{sec:latency}

We measure attention-only latency from 64k to 512k tokens to isolate the
$O(n)$ scaling behaviour predicted by the mass cliff analysis.
Table~\ref{tab:latency} and Figure~\ref{fig:latency} present three CRISP configurations against both FlexPrefill baselines.
All three start near latency parity with their FlexPrefill counterparts at short contexts and pull ahead as sequence length grows.

\textbf{CRISP $\alpha{=}1.5$ (speed-focused)} is already 7\% faster than FP $\gamma{=}0.95$ at 64k and extends to 18\% faster at 512k, reaching $5.40\times$ speedup over FlashAttention---the fastest configuration overall while still matching FP $\gamma{=}0.95$ on aggregate accuracy (see Appendix~\ref{app:hyperparams}).

\textbf{CRISP $\alpha{=}1.25$ (balanced)} tracks FP $\gamma{=}0.95$ closely at short contexts and pulls ahead by 13--17\% at 262k--512k, reaching $5.30\times$ at 512k vs.\ $4.41\times$ for FP $\gamma{=}0.95$.

\textbf{CRISP $\alpha{=}1.0$ (accuracy-focused)} is latency-matched to FP $\gamma{=}0.97$ at short contexts but 25\% faster at 512k, reaching $5.17\times$ vs.\ $3.90\times$---while delivering the highest accuracy across benchmarks.

The widening gap as context grows is the $O(n)$ theorem made concrete: noise accumulation is
$O(n)$, CRISP eliminates it, and the efficiency advantage compounds accordingly.

\paragraph{Short contexts.}
Dynamic sparse attention is not profitable at every length.
Routing is a fixed per-head cost while attention compute grows quadratically, so at short contexts it dominates what it saves.
On Llama-3.1-8B at 8K, FlexPrefill runs at $225.9$\,ms and CRISP at $209.9$\,ms, $4.4\times$ and $4.1\times$ the cost of dense attention respectively; by 64k CRISP is already $1.76\times$ faster than dense, so the crossover falls between these two points.
The ordering at 8K is informative: CRISP is $7\%$ faster than FlexPrefill precisely here, because replacing the pooled matmul and KL divergence with a constant-time reduction matters most where routing, not attention, is the dominant cost.
Deployments serving mostly short prompts should gate sparse prefilling on a context-length threshold.

\begin{table}[h]
\centering
\small
\resizebox{\columnwidth}{!}{%
\begin{tabular}{lcccc}
\toprule
\textbf{Method} & \textbf{64k} & \textbf{131k} & \textbf{262k} & \textbf{512k} \\
\midrule
FP $\gamma{=}0.95$    & 1{,}547ms & 4{,}118ms & 13{,}428ms & 40{,}949ms \\
FP $\gamma{=}0.97$    & 1{,}690ms & 5{,}001ms & 13{,}413ms & 46{,}369ms \\
\midrule
CRISP $\alpha{=}1.5$  & \textbf{1{,}441ms} \scriptsize{(\textbf{--7\%})} & \textbf{3{,}940ms} \scriptsize{(\textbf{--4\%})} & \textbf{10{,}982ms} \scriptsize{(\textbf{--18\%})} & \textbf{33{,}454ms} \scriptsize{(\textbf{--18\%})} \\
CRISP $\alpha{=}1.25$ & 1{,}589ms \scriptsize{(+3\%)} & 4{,}314ms \scriptsize{(+5\%)} & 11{,}714ms \scriptsize{(--13\%)} & 34{,}111ms \scriptsize{(--17\%)} \\
CRISP $\alpha{=}1.0$  & 1{,}776ms \scriptsize{(+5\%)} & 4{,}801ms \scriptsize{(--4\%)} & 12{,}740ms \scriptsize{(--5\%)} & 34{,}983ms \scriptsize{(--25\%)} \\
\bottomrule
\end{tabular}%
}
\caption{Attention-only latency (mean, single H100).
FlexPrefill baselines are grouped at top; CRISP configurations below, ordered by speed.
All CRISP configurations overtake FlexPrefill at long contexts, with the efficiency advantage widening as sequence length grows---consistent with the $O(n)$ noise elimination predicted in \S\ref{sec:failure2}.
Overhead/saving in parentheses relative to FP $\gamma{=}0.95$ for $\alpha{\in}\{1.5, 1.25\}$ and FP $\gamma{=}0.97$ for $\alpha{=}1.0$.}
\label{tab:latency}
\end{table}

\subsection{Ablation: Component Contributions}
\label{sec:ablation}
Table~\ref{tab:ablation} isolates each component.
The two are not independent: Abl.~1 alone \emph{hurts} Llama-3.1-8B ($-2.2$\,pp IB, $-1.4$\,pp RULER): by routing more heads to \textsc{vs} than JSD does, it exposes precisely those additional heads to the $\gamma$-cumsum selection that the mass cliff defeats.
Abl.~2 is broadly beneficial on its own, and full CRISP recovers the regression Abl.~1 introduces: the two limitations interact, and neither fix alone yields stable long-context behaviour across both architectures.

\begin{table}[h]
\centering
\small
\resizebox{\columnwidth}{!}{%
\begin{tabular}{lcccccc}
\toprule
& \multicolumn{3}{c}{\textbf{Llama-3.1-8B}} & \multicolumn{3}{c}{\textbf{Qwen2.5-7B}} \\
\cmidrule(lr){2-4} \cmidrule(lr){5-7}
\textbf{Method} & \textbf{IB} & \textbf{LB} & \textbf{RULER}
                & \textbf{IB} & \textbf{LB} & \textbf{RULER} \\
\midrule
FlexPrefill ($\gamma{=}0.95$) & 47.4 & 46.92 & \textbf{89.22} & 25.5 & 45.57 & 75.52 \\
\midrule
Abl.~1: $C_{\text{struct}}$ only  & 45.2 & 47.14 & 87.82 & \textbf{29.7} & 45.93 & 74.92 \\
Abl.~2: sink-aware only           & \textbf{48.7} & \underline{47.70} & \underline{89.10} & 25.7 & \underline{47.10} & \textbf{77.08} \\
\midrule
\textbf{CRISP} ($\tau{=}0.2, \alpha{=}1.0$) & \textbf{48.7} & \textbf{47.77} & 88.82 & \underline{28.7} & 
\textbf{47.23} & \underline{76.20} \\
\bottomrule
\end{tabular}%
}
\caption{Component ablation. Abl.~1 isolates $C_{\text{struct}}$ routing with
$\gamma$-cumsum VS selection; Abl.~2 isolates sink-aware selection with JSD
routing.
Full CRISP demonstrates the most robust performance across both models, 
recovering from regressions caused by partial systems.}
\label{tab:ablation}
\end{table}

%───────────────────────────────────────────────────────────────────────────────
\section{Related Work}
\label{sec:related}

\paragraph{Fixed and Offline Sparse Attention.}
Early sparse attention methods employ fixed structural patterns, including
Sparse Transformers~\citep{child2019generating},
global-local combinations as in BigBird~\citep{zaheer2020bigbird} and Longformer~\citep{beltagy2020longformer}, sliding
window attention~\citep{jiang2023mistral7b}, and the attention sink mechanism of
StreamingLLM~\citep{xiao2024efficient}, which combines initial-token sinks
with a local sliding window.
MInference~\citep{jiang2024minference} advances
beyond purely fixed patterns by profiling per-head sparse patterns offline and
generating dynamic indices at runtime within those profiles.
However, offline
pattern assignment cannot adapt to input-dependent variation in sink mass: when
mass concentration changes across inputs or context lengths, a fixed profile
either over-selects sinks or under-selects signal.
This limitation is
quantified in our experiments, where MInference trails FlexPrefill by 5--8~pp
on LongBench and collapses on long-context RULER---constituting empirical
evidence that the mass cliff requires \emph{online} navigation.

\paragraph{Dynamic Sparse Attention.}
FlexPrefill~\citep{lai2025flexprefill} represents the current state of the art
in dynamic sparse attention by routing heads and allocating budgets at prefill
time based on each input's attention structure.
Other recent approaches dynamically select tokens by tracking ``heavy hitters'' during generation \citep{zhang2023h2o}, utilize query-aware page routing \citep{tang2024questqueryawaresparsityefficient}, route heads to a mixture of sparse attention patterns \citep{fu2025mixtureattentionspansoptimizing}, approximate attention scores via query-component slicing \citep{ribar2024sparq}, or employ locality-sensitive hashing for importance sampling \citep{chen2025magicpig}.
Concurrently, Twilight~\citep{lin2025twilight} replaces fixed top-$k$ budgets with top-$p$ pruning that adaptively selects the minimum token set whose attention weights exceed a cumulative threshold, addressing the over/under-selection problem inherent in static budgets during decoding.
Its scope is complementary to ours rather than overlapping: Twilight operates on a single query row at decode time and targets KV-cache memory bandwidth, whereas CRISP operates on the full prefill attention map and targets a compute-bound cost, so the two could in principle be composed (CRISP for prefilling, Twilight for decoding).
Separately, \citet{nawrot2026sparsefrontiersparseattention} provide a large-scale isoCost analysis showing that sparse attention efficiency varies dramatically across task types, with aggregation and multi-hop tasks degrading at sparsity levels that retrieval tasks tolerate---consistent with our finding that cumulative thresholding fails to preserve task-relevant signal at scale.

\paragraph{Subquadratic Alternatives.}
A parallel line of work removes the quadratic cost rather than sparsifying it, replacing softmax attention with linear or recurrent formulations: state-space models such as Mamba~\citep{gu2024mamba}, gated linear attention such as Gated DeltaNet~\citep{yang2025gateddelta}, and post-hoc linearization of pretrained Transformers, as in LoLCATs~\citep{zhang2025lolcats} and Lizard~\citep{nguyen2026lizard}.
These change the architecture and usually require distillation or fine-tuning, whereas CRISP is training-free and computes exact softmax attention on the blocks it selects; the two directions are complementary.

Theoretical grounding for the
routing dichotomy between structured and diverse attention heads has remained
limited: \citet{likhosherstov2021expressivepowerselfattentionmatrices} implicitly connected attention
concentration to representational capacity, and \citet{zhang-etal-2025-attention} established attention entropy as a key factor governing parallel context encoding efficiency, but without a formal account of the
VS/PE dichotomy or its interaction with cumulative thresholding.
The present work provides that foundation, identifying the mass cliff as an intrinsic structural consequence of autoregressive softmax attention and proving that strictly cumulative coverage criterion \citep{lai2025flexprefill} is asymptotically ill-suited to ultra-long contexts.

%───────────────────────────────────────────────────────────────────────────────
\section{Conclusion}
\label{sec:conclusion}
We identified and resolved two principled limitations in dynamic sparse attention.
First, we replaced indirect JSD-based routing with $C_{\text{struct}}$, a structural-mass proxy that reproduces JSD's routing decisions while eliminating the pooling matmul and KL divergence overhead.
Second, we formalised the post-softmax \emph{mass cliff}, proving that cumulative $\gamma$-thresholding inherently accumulates $O(n)$ background noise.
By implementing a sink-aware noise-floor threshold, CRISP explicitly separates signal from architectural noise.
Consequently, CRISP establishes a new performance hierarchy, recovering up to +28.0~pp on retrieval tasks, achieving parity with exact dense attention on InfiniteBench, and delivering up to a 5.30$\times$ latency speedup at 512k tokens.
Ultimately, navigating the mass cliff provides a robust, theoretically grounded foundation for scaling long-context sparse attention.

\section*{Limitations}
\textbf{Architectural Scope.} $C_{\text{struct}}$ is a structural measurement, not a formal entropy quantity, and it works as a routing signal because low-entropy heads in current sink-having architectures concentrate their mass at sinks and recency (Table~\ref{tab:routing_analysis}).
This is an empirical property of the models we study, not a guarantee.
Architectures that suppress attention sinks---for instance via gated attention---break that correspondence, and CRISP's assumptions do not hold there.
Sparse attention in the sink-free regime is largely unexplored, and the prior methods we compare against \citep{jiang2024minference, lai2025flexprefill} do not address it either;
the entropy argument for the \textsc{vs}/\textsc{pe} dichotomy itself (Appendix~\ref{app:proof_coverage}) is a property of softmax attention and would still apply in principle, but a different structural proxy would be required.
Our evidence spans two model families at 7--8B scale, which bounds how far these regularities should be assumed to generalise.

\textbf{Hybrid Heads.} CRISP inherits FlexPrefill's binary \textsc{vs}/\textsc{pe} routing.
Real heads are not always cleanly one or the other: some carry mixed structure, and forcing such a head onto the \textsc{vs} path lets the sink-aware threshold discard tokens that do not fit that pattern.
Softer schemes---probabilistic mixing, or per-head confidence-weighted decisions---are left to future work.

\textbf{Scaling and Decoding.} Empirical verification on models $>8$B or contexts beyond established benchmarks remains future work.
Additionally, this addresses prefilling only; extension to token-level decoding under causal masking requires characterizing the mass cliff dynamically during auto-regressive generation.

\textbf{Path Selection.} CRISP improvements focus heavily on the VS path.
While PE heads lack a sharp mass cliff hierarchy, full improvements to the PE path and dynamic scheduling of $\alpha$ to mitigate precision-coverage tradeoffs in aggregation tasks (e.g., Llama RULER) are left to future treatment.

\bibliography{custom}

%───────────────────────────────────────────────────────────────────────────────
\appendix

\section{FlexPrefill Baseline Algorithms}
\label{app:flex_algo}

For completeness, we provide the algorithmic details of the FlexPrefill baseline \citep{lai2025flexprefill}.
Algorithm~\ref{alg:flex_routing} describes the JSD-based routing mechanism, and Algorithm~\ref{alg:flex_vs} describes the dual-direction cumulative thresholding used for VS heads.

\begin{algorithm}[H]
\caption{FlexPrefill Pattern Search (JSD)}
\label{alg:flex_routing}
\begin{algorithmic}[1]
\State \textbf{Input:} $Q, K, \tau$
\State $\hat{Q} \leftarrow Q[-b:]$ \Comment{Representative query block}
\State $\bar{a} \leftarrow \mathrm{Softmax}(\mathrm{pool}(\hat{Q})\mathrm{pool}(K)^\top/\sqrt{d})$ \Comment{Pooled estimate}
\State $\hat{a} \leftarrow \mathrm{sumpool}(\mathrm{Softmax}(\hat{Q}K^\top/\sqrt{d}))$ \Comment{Per-query mean}
\State $D_{JS} \leftarrow \sqrt{\tfrac{1}{2}KL(\hat{a}||m) + \tfrac{1}{2}KL(\bar{a}||m)}$ \textbf{where} $m = \tfrac{1}{2}(\hat{a} + \bar{a})$
\If{$D_{JS} \geq \tau$} 
    \State \Return \textsc{vs} 
\Else 
    \State \Return \textsc{pe} 
\EndIf
\end{algorithmic}
\end{algorithm}

\begin{algorithm}[H]
\caption{FlexPrefill VS Index Selection ($\gamma$)}
\label{alg:flex_vs}
\begin{algorithmic}[1]
\State \textbf{Input:} $\hat{A}$ (proxy attention), $\gamma$
\State $a_v \leftarrow \mathrm{sum}(\hat{A}, \text{axis=0})\,/\!\sum_{i,j}\hat{A}[i,j]$ \Comment{Normalized vertical scores}
\State $a_s \leftarrow \mathrm{sum\_slash}(\hat{A})\,/\!\sum_{i,j}\hat{A}[i,j]$ \Comment{Normalized diagonal scores}
\State $S_v \leftarrow \{j : \sum \mathrm{sort}(a_v) \geq \gamma\}$ \Comment{Vertical cumulative set}
\State $S_s \leftarrow \{j : \sum \mathrm{sort}(a_s) \geq \gamma\}$ \Comment{Slash cumulative set}
\State \Return $S_v \cup 
S_s \cup \{0, N_b-1\}$ \Comment{Union with sinks and local}
\end{algorithmic}
\end{algorithm}

\paragraph{Comparison with CRISP.} CRISP replaces Algorithm~\ref{alg:flex_routing} with a constant-time indexed slice reduction over $\hat{A}$ (eliminating the $O(nd/b)$ pooled matmul and $O(n/b)$ KL divergence overhead in Lines 3-5), and replaces Algorithm~\ref{alg:flex_vs} with block-level sink-aware thresholding (eliminating the dual-direction decomposition in Lines 2--5).

\section{Entropy-Coverage Relationship}
\label{app:proof_coverage}

We provide the information-theoretic intuition underlying the 
VS/PE routing dichotomy.

\textbf{Effective support size.}
The exponentiated entropy $e^{H(a)}$ is the standard information-theoretic
measure of the \emph{effective support size} of a distribution $a$
\citep{Campbell1966}.
At the two extremes:
\begin{itemize}
    \item \textbf{Concentrated:}
    $H(a) \approx 0 \Rightarrow e^{H(a)} \approx 1$.
    A single element holds most mass; coverage $\gamma$ is achieved
    with very few tokens.
    \item \textbf{Uniform:}
    $H(a) = \log n \Rightarrow e^{H(a)} = n$.
    Mass is spread evenly; coverage $\gamma$ requires $\gamma n$ tokens.
\end{itemize}

\textbf{Monotone scaling.}
Between these extremes, concentrating mass on fewer tokens 
(lowering entropy) reduces the number of tokens needed 
for coverage $\gamma$, so low-entropy heads require compact 
index sets while high-entropy heads require broad coverage.

\textbf{Connection to routing.}
For low-entropy heads ($e^{H(a)} \ll n$), a compact sparse index set
suffices. Prior work has consistently observed that in autoregressive 
transformers, low-entropy heads concentrate mass at structurally 
predictable positions---sinks, vertical columns, and slash 
diagonals \citep{xiao2024efficient, jiang2024minference, lai2025flexprefill}---making 
the \textsc{vs} path a natural fit.
For high-entropy heads ($e^{H(a)} \approx n$), the required
support is large and no compact structural pattern dominates; 
pooled estimation over the full sequence is more appropriate.
This is the basis for the routing dichotomy.

\section{Mass Cliff: Quantitative Analysis}
\label{app:cliff}

For a VS head with $k$ signal tokens (sinks and task-relevant) 
at pre-softmax logit $z$ and $n - k$ background tokens at 
logit $z - \Delta$, softmax gives each background token mass:
\[
\epsilon = \frac{e^{-\Delta}}{k + (n-k) \cdot e^{-\Delta}}
\]
For long contexts where $n \gg k \cdot e^{\Delta}$, the 
denominator is dominated by $(n-k) \cdot e^{-\Delta}$, yielding 
$\epsilon \approx 1/n$. The garbage token count required to 
satisfy residual threshold $\gamma_{\text{residual}} = \gamma - 
a_{\text{sink}} - a_{\text{important}}$ is then:
\[
N_{\text{garbage}} = \gamma_{\text{residual}} / \epsilon 
\approx \gamma_{\text{residual}} \cdot n
\]
demonstrating $O(n)$ scaling at long contexts. This is 
consistent with the general sparsity framework of 
\citet{deng2024attention}.

\section{Additional Mass Cliff Examples}
\label{app:extra_cliffs}

\begin{figure*}[h]
  \centering
  \includegraphics[width=\linewidth]{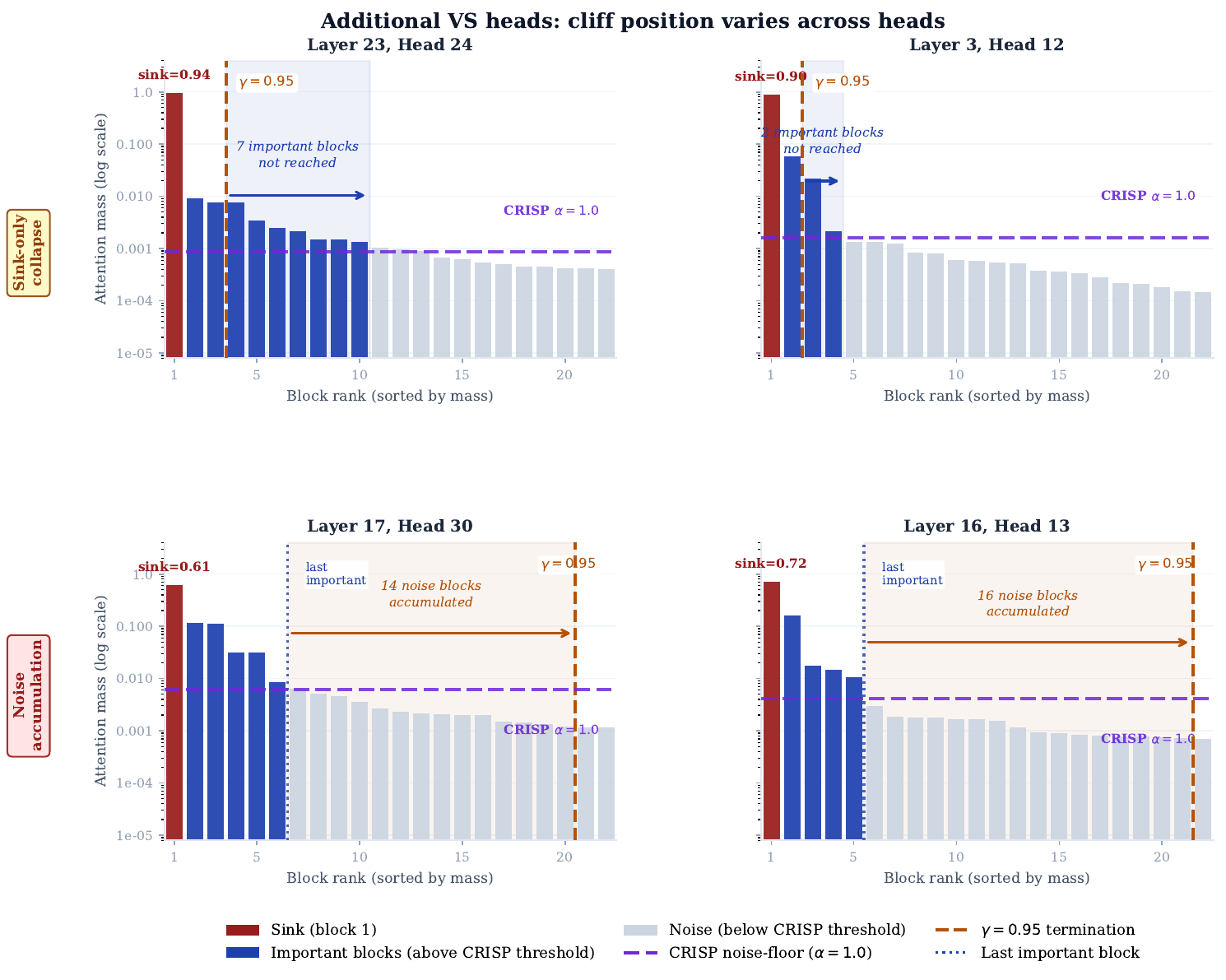}
  \caption{Extended visualization of the post-softmax attention mass distribution across four additional representative VS heads.
Both failure modes (sink-only collapse and noise accumulation) arise from the same fixed threshold applied to heads whose cliff sits at different positions, indicating that the failure is a structural limitation of cumulative thresholding rather than a calibration artefact.
These panels show cross-head heterogeneity at a fixed input; we do not claim from them that a single head's cliff moves across inputs.}
  \label{fig:cliff_histogram_full}
\end{figure*}

Figure~\ref{fig:cliff_histogram_full} provides additional visual examples of the post-softmax mass cliff across four entirely different layers and heads of Llama-3.1-8B.
These further demonstrate the two structural failure modes of cumulative $\gamma$-thresholding.
Wherever the cliff falls for a given head, the CRISP noise-floor criterion ($\alpha{=}1.0$) isolates the task-relevant signal without accumulating $O(n)$ background noise.

\section{Hyperparameter Robustness}
\label{app:hyperparams}

Tables~\ref{tab:alpha_sweep} and~\ref{tab:tau_sweep} report sensitivity to
the two CRISP hyperparameters, $\alpha$ and $\tau_{\text{proxy}}$, on
Llama-3.1-8B and Qwen2.5-7B respectively.
Performance degrades monotonically as $\alpha$ increases above 1.0, 
confirming that raising the threshold past the noise floor progressively 
discards signal.
Notably, $\alpha{=}0.75$ performs comparably to $\alpha{=}1.0$, 
matching or exceeding it on some benchmarks (e.g., Qwen LongBench). 
This is expected: a lower $\alpha$ retains more blocks, benefiting tasks 
that require broader coverage at the cost of admitting some near-threshold 
noise. The noise floor is thus better understood as a stable region 
rather than a knife-edge---values in $[0.75, 1.0]$ all correctly separate 
signal from deep background noise, differing only in how aggressively 
they treat borderline blocks. We recommend $\alpha{=}1.0$ as the default 
because it is the only value with a calibration-free interpretation 
(above-average residual mass), removing the need for per-task tuning.
Even at $\alpha{=}1.5$---well beyond the principled default---CRISP still Pareto-dominates FlexPrefill $\gamma{=}0.95$ on aggregate accuracy while running faster (Figure~\ref{fig:results_overview}), demonstrating graceful degradation.

Critically, $\tau_{\text{proxy}}$ sensitivity is low: varying
$\tau_{\text{proxy}} \in \{0.15, 0.2, 0.25\}$ produces a spread of less than
0.7pp on InfiniteBench and less than 1.2pp on LongBench across both model
families, with no single value dominating uniformly.
This insensitivity
confirms that $C_{\text{struct}}$ provides a stable routing signal across a
wide threshold range, requiring no careful per-model calibration.

\begin{table}[h]
\centering
\small
\resizebox{\columnwidth}{!}{%
\begin{tabular}{lcccccc}
\toprule
& \multicolumn{3}{c}{\textbf{Llama-3.1-8B}} & \multicolumn{3}{c}{\textbf{Qwen2.5-7B}} \\
\cmidrule(lr){2-4} \cmidrule(lr){5-7}
$\alpha$ & \textbf{IB} & \textbf{LB} & \textbf{RULER}
                       & \textbf{IB} & \textbf{LB} & \textbf{RULER} \\
\midrule
0.75 & \textbf{48.65} & \underline{47.71} & 88.54 & \textbf{28.96} & \textbf{47.76} & \underline{75.96} \\
\textbf{1.0}  & \textbf{48.65} & \textbf{47.77} & \textbf{88.82} & 28.73 & \underline{47.23} & \textbf{76.20} \\
1.25 & \underline{48.03} & 46.92 & \underline{88.62} & 28.35 & 46.76 & 75.80 \\
1.5  & 47.41 & 46.03 & 87.94 & \underline{28.92} & 46.18 & 75.48 \\
\bottomrule
\end{tabular}%
}
\caption{$\alpha$ sensitivity ($\tau_{\text{proxy}}{=}0.2$).
Values in $[0.75, 1.0]$ perform comparably, with degradation above 1.0 as signal blocks are progressively discarded.
We recommend $\alpha{=}1.0$ as a calibration-free default.
Even at $\alpha{=}1.5$, CRISP Pareto-dominates FlexPrefill $\gamma{=}0.95$ on aggregate accuracy (55.3 vs 55.0, averaging IB, RULER, and LB across both models) while running 21\% faster at 131k tokens.}
\label{tab:alpha_sweep}
\end{table}

\begin{table}[h]
\centering
\small
\resizebox{\columnwidth}{!}{%
\begin{tabular}{lcccccc}
\toprule
& \multicolumn{3}{c}{\textbf{Llama-3.1-8B}} & \multicolumn{3}{c}{\textbf{Qwen2.5-7B}} \\
\cmidrule(lr){2-4} \cmidrule(lr){5-7}
$\tau_{\text{proxy}}$ & \textbf{IB} & \textbf{LB} & \textbf{RULER}
                      & \textbf{IB} & \textbf{LB} & \textbf{RULER} \\
\midrule
0.15 & \textbf{48.72} & \textbf{47.93} & \textbf{88.90} & 27.12 & \textbf{47.32} & \underline{76.08} \\
\textbf{0.2}  & \underline{48.65} & \underline{47.77} & \underline{88.82} & \underline{28.73} & \underline{47.23} & \textbf{76.20} 
\\
0.25 & 48.09 & 47.71 & 88.70 & \textbf{29.23} & 46.77 & 75.64 \\
\midrule
Spread & 0.60 & 1.22pp & 0.20pp & 2.10 & 0.55pp & 0.56pp \\
\bottomrule
\end{tabular}%
}
\caption{$\tau_{\text{proxy}}$ sensitivity ($\alpha{=}1.0$).
No single value
dominates uniformly across benchmarks and models; the spread across the full
range is $\leq$1.2pp, confirming that $C_{\text{struct}}$ routing is robust
to threshold choice and requires no per-model calibration.}
\label{tab:tau_sweep}
\end{table}

\section{Why the PE Path Keeps $\gamma$-cumsum}
\label{app:pe_heads}

\begin{figure*}[t]
  \centering
  \includegraphics[width=\linewidth]{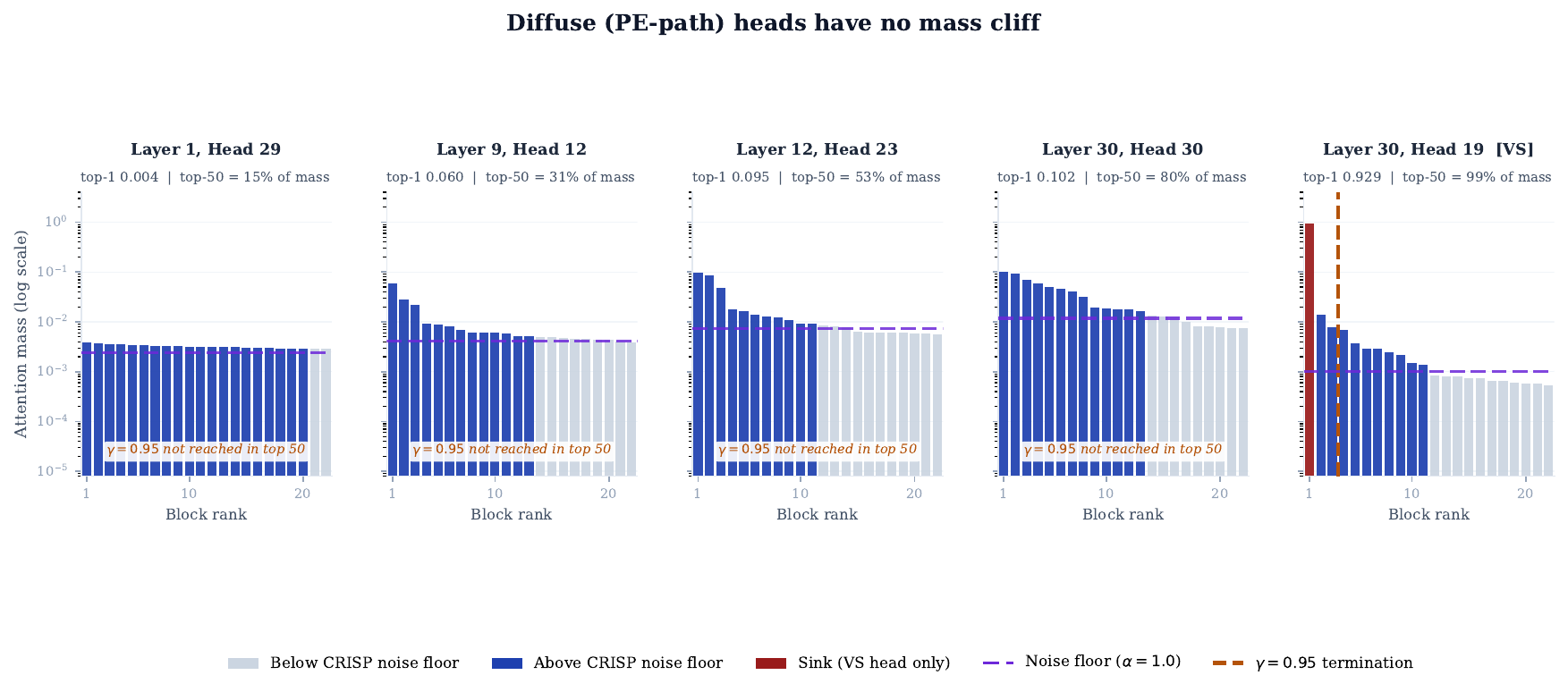}
  \caption{Block-mass profiles of four diffuse (PE-path) heads from different layers of Llama-3.1-8B, with one \textsc{vs} head at right for scale.
  The \textsc{vs} head shows the sink spike and the sharp cliff of Figure~\ref{fig:cliff_histogram};
  the diffuse heads show neither, and $\gamma{=}0.95$ is not reached within the top 50 blocks.
  Heads are identified by top-1 block mass, not by recomputing the routing decision (see text).}
  \label{fig:pe_heads}
\end{figure*}

\S\ref{sec:sink_aware_fix} argues that PE heads do not exhibit the mass cliff and can therefore retain $\gamma$-cumsum.
Figure~\ref{fig:pe_heads} supports that empirically, using the same Llama-3.1-8B block-mass dump that produces Figures~\ref{fig:cliff_histogram} and~\ref{fig:cliff_histogram_full}.

We identify diffuse heads by their top-1 block mass, drawing four from different layers.
Their profiles decay gently and monotonically with no boundary anywhere: the noise floor sits \emph{inside} the bulk of the distribution rather than at a step, which is exactly the regime in which a mean-based threshold has nothing to separate.
The concentration statistic makes the same point numerically---the top-50 blocks hold only $15\%$ to $80\%$ of these heads' mass, against $99\%$ for the \textsc{vs} head shown for contrast, whose sink block alone carries $0.929$.
Cumulative thresholding does not over-collect here because there is no near-zero background band to fall into: $\gamma{=}0.95$ is not even reached within the top 50 blocks.
Accumulating to coverage therefore selects a representative subset rather than a disproportionate residual, which is why replacing it would gain nothing on this path.

Two caveats.
The dump stores each head's top-50 block masses sorted by magnitude, so it characterises \emph{concentration} and does not let us recompute the routing decision itself;
we identify these heads as diffuse, and rely on the measured $91.3\%$ rate at which top-1 mass sits at an anchor position (Table~\ref{tab:routing_analysis}) to connect low concentration to a low $C_{\text{struct}}$ and hence to the PE path.
The dump also covers a single model and input, so this establishes the shape of the PE-head distribution, not its variability.

\section{Per-Benchmark Pareto Plots}
\label{app:per_bench_pareto}

\begin{figure*}[t]
  \centering
  \includegraphics[width=\linewidth]{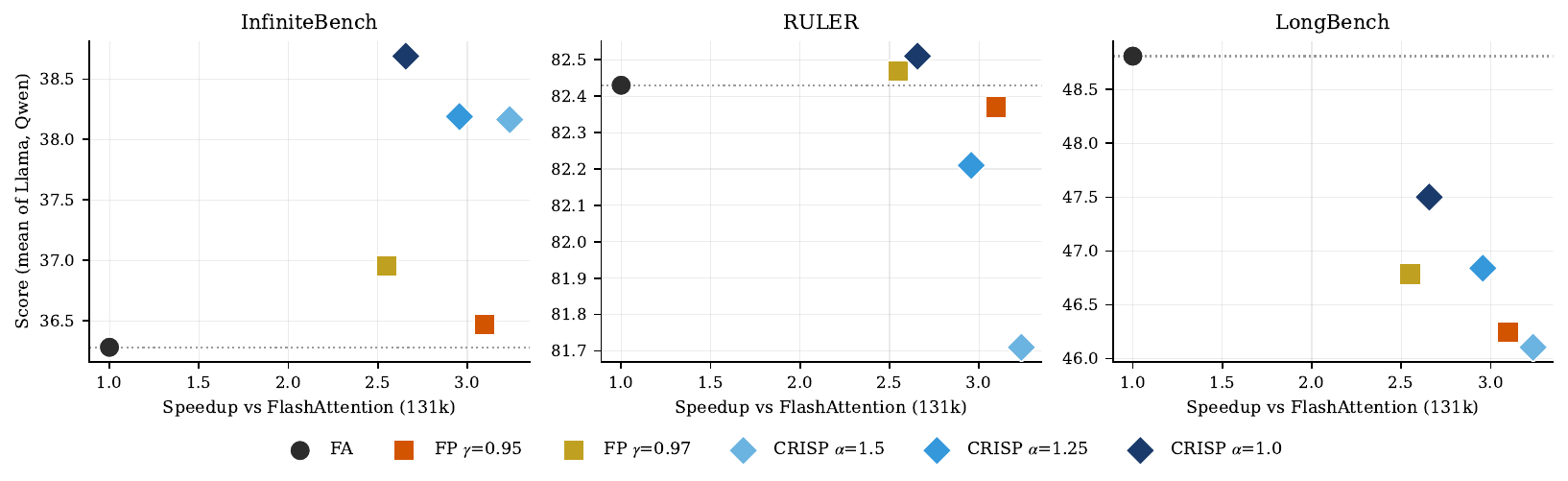}
  \caption{Accuracy-efficiency tradeoff per benchmark at 131k tokens, scores averaged over Llama-3.1-8B and Qwen2.5-7B.
  Dotted line marks dense attention. The aggregate view of Figure~\ref{fig:results_overview}(a) masks the RULER result, which is shown here directly.}
  \label{fig:pareto_per_bench}
\end{figure*}

Figure~\ref{fig:results_overview}(a) aggregates three benchmarks and two models into one point per method, which hides task-specific behaviour.
Figure~\ref{fig:pareto_per_bench} separates them.
CRISP $\alpha{=}1.0$ leads on InfiniteBench and LongBench, but on RULER it sits marginally below both FlexPrefill settings on the Llama side---the precision-coverage tradeoff discussed in \S\ref{sec:general}.
We show it here rather than leaving it inside an average.

\section{Additional Model and Variant Checks}
\label{app:extra_checks}

\paragraph{Transfer to a newer architecture.}
Table~\ref{tab:qwen3} reports a check on Qwen3-4B-Instruct-2507 under the same configuration ($\gamma{=}0.95$, $\tau_{\text{proxy}}{=}0.2$, $B{=}128$, $\alpha{=}1.0$).
CRISP exceeds dense attention on InfiniteBench by $+2.6$\,pp, driven by retrieval (kv\_retrieval $0.220$ vs $0.128$; passkey $0.542$ vs $0.468$), comes within $0.30$\,pp of dense on LongBench---the smallest gap of the sparse methods---and stays within $\pm1.7$\,pp on RULER across all six context lengths.
The shape of the result matches Qwen2.5-7B, which suggests the structural mechanism is not specific to the two models of \S\ref{sec:experiments}.
This is a single-scale check, not a substitute for a full sweep.

\begin{table}[h]
\centering
\small
\begin{tabular}{lccc}
\toprule
\textbf{Benchmark} & \textbf{FA} & \textbf{FP} & \textbf{CRISP} \\
\midrule
InfiniteBench        & 28.67 & 29.34 & \textbf{31.28} \\
LongBench            & \textbf{46.64} & 44.99 & \underline{46.34} \\
RULER (4K--131K)     & \textbf{87.50} & \underline{87.14} & 86.82 \\
\bottomrule
\end{tabular}
\caption{Qwen3-4B-Instruct-2507. FA = FlashAttention, FP = FlexPrefill $\gamma{=}0.95$, CRISP at $\alpha{=}1.0$.}
\label{tab:qwen3}
\end{table}

\paragraph{Median instead of mean as the noise floor.}
The floor $\mu$ is a mean over non-retained blocks, so a single dominant block can pull it above weaker but genuine signal.
A median-based floor is the natural robust alternative, and Table~\ref{tab:median} compares the two under an otherwise identical configuration.
The two are within $\pm0.7$\,pp of each other in both directions on every benchmark, with neither dominating.
The clumpy-noise failure mode is therefore real in principle but does not drive end-task accuracy on these workloads;
we keep the mean because it is what makes $\alpha{=}1.0$ interpretable as ``above-average residual mass''.

\begin{table}[h]
\centering
\small
\begin{tabular}{llccc}
\toprule
\textbf{Model} & \textbf{Floor} & \textbf{IB} & \textbf{LB} & \textbf{RULER} \\
\midrule
\multirow{2}{*}{Llama-3.1-8B} & mean   & 48.65 & \textbf{47.77} & 88.82 \\
                              & median & \textbf{48.81} & 47.52 & \textbf{88.86} \\
\midrule
\multirow{2}{*}{Qwen2.5-7B}   & mean   & \textbf{28.73} & 47.23 & \textbf{76.20} \\
                              & median & 28.68 & \textbf{47.91} & 75.56 \\
\bottomrule
\end{tabular}
\caption{Mean vs.\ median noise floor ($\gamma{=}0.95$, $\alpha{=}1.0$, $\tau_{\text{proxy}}{=}0.2$).}
\label{tab:median}
\end{table}

\section{Full Benchmark Results}
\label{app:full_results}

\begin{table*}[h]
\centering
\small
\resizebox{\textwidth}{!}{%
\begin{tabular}{lcccccccccccc}
\toprule
\multirow{2}{*}{\textbf{Task}}
    & \multicolumn{6}{c}{\textbf{Llama-3.1-8B}}
    & \multicolumn{6}{c}{\textbf{Qwen2.5-7B}} \\
\cmidrule(lr){2-7} \cmidrule(lr){8-13}
& \textbf{FA} & \textbf{MInf} & \textbf{FP 0.95} & \textbf{FP 0.97} & \textbf{CRISP} & \textbf{$\Delta$vs FP}
& \textbf{FA} & \textbf{MInf} & \textbf{FP 0.95} & \textbf{FP 0.97} & \textbf{CRISP} & \textbf{$\Delta$vs FP} \\
\midrule
kv\_retrieval         & 62.6 & 21.0 & 43.8 & \underline{48.4} & \textbf{61.6} & +17.8pp & 0.0 & 0.0 & \textbf{0.4} & 0.0 & 
0.0 & -- \\
math\_find            & 25.1 & 25.4 & \textbf{30.9} & 25.7 & \underline{30.3} & --0.6pp & 33.7 & 37.7 & \textbf{39.7} & \underline{39.4} & 39.1 & --0.6pp \\
longbook\_sum\_eng    & 32.3 & 30.9 & 31.7 & \textbf{32.4} & \underline{32.3} & +0.6pp  & 18.3 & 16.1 & 18.4 & \underline{18.5} & \textbf{18.6} & +0.2pp \\
longbook\_qa\_eng     & 26.0 & 22.7 & 25.3 & \underline{25.8} & \textbf{26.2} & +0.9pp  & 5.2 & 4.8 & \underline{5.2} & 5.1 & \textbf{5.5} & +0.3pp \\
longbook\_choice\_eng & 
69.4 & 65.1 & 69.4 & \textbf{70.7} & \underline{69.9} & +0.5pp  & 44.1 & 38.0 & \textbf{41.1} & \underline{40.6} & \textbf{41.1} & 0.0pp \\
longdialogue\_qa\_eng & 20.5 & 14.0 & \underline{19.0} & 18.0 & \textbf{21.5} & +2.5pp  & 15.5 & 10.0 & \textbf{15.0} & 11.0 & \underline{12.0} & --3.0pp \\
longbook\_qa\_chn     & 32.7 & 27.7 & \textbf{35.0} & \textbf{35.0} & \underline{31.9} & --3.1pp & 9.5 & \textbf{9.2} & 8.6 & 8.4 & \underline{9.0} & +0.4pp \\
code\_debug           & 18.0 & \textbf{21.3} & \underline{20.6} & 19.0 & 17.8 
& --2.8pp & 19.3 & 16.0 & 17.5 & \textbf{19.3} & \underline{17.8} & +0.3pp \\
passkey               & 99.3 & 59.8 & \underline{98.6} & 97.6 & \textbf{98.8} & +0.2pp  & 0.3 & \underline{41.4} & 22.4 & 37.0 & \textbf{50.3} & +27.9pp \\
number\_string        & 99.7 & 76.4 & \underline{99.7} & \textbf{99.8} & \textbf{99.8} & +0.1pp  & 94.1 & 78.0 & \underline{87.3} & \underline{87.3} & \textbf{93.9} & +6.6pp \\
\midrule
\textbf{Mean} & 48.6 & 36.5 & \underline{47.4} & 47.3 & \textbf{48.7} & \textbf{+1.3pp}
 
             & 24.0 & 25.1 & 25.5 & \underline{26.7} & \textbf{28.7} & \textbf{+3.2pp} \\
\bottomrule
\end{tabular}%
}
\caption{Full InfiniteBench results.
FA = FlashAttention; MInf = MInference; FP = FlexPrefill.
CRISP $\alpha{=}1.0$ achieves parity with exact dense attention on both models.
$\Delta$ is vs FP 0.95.}
\label{tab:infinitebench}
\end{table*}

\begin{table*}[h]
\centering
\small
\resizebox{\textwidth}{!}{%
\begin{tabular}{lcccccccccccc}
\toprule
\multirow{2}{*}{\textbf{Task}}
    & \multicolumn{6}{c}{\textbf{Llama-3.1-8B}}
    & \multicolumn{6}{c}{\textbf{Qwen2.5-7B}} \\
\cmidrule(lr){2-7} \cmidrule(lr){8-13}
& \textbf{FA} & \textbf{MInf} & \textbf{FP 0.95} & \textbf{FP 0.97} & \textbf{CRISP} & \textbf{$\Delta$vs FP}
& \textbf{FA} & \textbf{MInf} & \textbf{FP 0.95} & \textbf{FP 0.97} & \textbf{CRISP} & \textbf{$\Delta$vs FP} \\
\midrule
narrativeqa            & 25.82 & \textbf{29.74} & 24.01 & 25.01 & \underline{25.76} & +1.75 & 25.47 & 11.87 & 23.88 & \textbf{25.32} & \underline{24.29} & +0.41 \\
qasper                 & 45.09 
& 26.63 & \underline{44.57} & 43.75 & \textbf{44.97} & +0.40 & 43.73 & 13.65 & \underline{41.69} & 41.61 & \textbf{43.76} & +2.07 \\
multifieldqa\_en       & 55.76 & 28.20 & 54.88 & \textbf{55.71} & \underline{55.25} & +0.37 & 52.42 & 32.47 & \underline{51.95} & \textbf{52.80} & 50.80 & --1.15 \\
multifieldqa\_zh       & 62.27 & \underline{61.58} & 61.01 & 61.23 & \textbf{62.15} & +1.14 & 61.92 & 43.24 & 61.96 & \textbf{62.26} & \underline{62.12} & +0.16 \\
hotpotqa               & 
56.43 & 17.37 & \underline{57.08} & 56.91 & \textbf{57.88} & +0.80 & 58.49 & 10.86 & 54.19 & \underline{55.98} & \textbf{58.59} & +4.40 \\
2wikimqa               & 45.38 & 15.77 & \underline{42.40} & \textbf{43.75} & 41.55 & --0.85 & 47.27 &  9.60 & 41.99 & \underline{42.28} & \textbf{45.06} & +3.07 \\
musique                & 31.73 & 10.76 & \textbf{32.33} & 31.82 & \underline{31.84} & --0.49 & 30.99 &  7.31 & \underline{31.67} & 30.49 
& \textbf{33.10} & +1.43 \\
dureader               & 33.71 & \textbf{35.06} & \underline{35.01} & 33.63 & 33.39 & --1.62 & 28.76 & \textbf{32.83} & 30.18 & 30.03 & \underline{30.55} & +0.37 \\
gov\_report            & 34.92 & 34.19 & \underline{34.83} & \textbf{34.91} & 34.46 & --0.37 & 31.97 & \textbf{32.22} & 31.83 & \underline{32.11} & 32.05 & +0.22 \\
qmsum                  & 24.48 & 
23.49 & \underline{24.91} & \textbf{25.01} & 24.78 & --0.13 & 23.16 & 20.84 & 22.86 & \underline{23.26} & \textbf{23.29} & +0.43 \\
multi\_news            & 27.24 & \underline{27.02} & \textbf{27.05} & 26.92 & 26.74 & --0.31 & 23.92 & 22.50 & \textbf{24.05} & \underline{23.88} & 23.85 & --0.20 \\
vcsum                  & 17.26 & 16.17 & 17.09 & \textbf{17.50} & \underline{17.47} & +0.38 & 16.19 & 15.77 & \underline{15.97} & \textbf{16.00} & 15.94 & --0.03 \\
samsum 
                & 43.93 & \underline{44.26} & 43.69 & 43.62 & \textbf{44.53} & +0.84 & 45.63 & 46.01 & \textbf{46.45} & 45.51 & \underline{46.05} & --0.40 \\
trec                   & 72.50 & \textbf{72.00} & \underline{70.50} & \underline{70.50} & 68.50 & --2.00 & 72.00 & \textbf{71.50} & \underline{70.50} & \textbf{71.50} & 68.00 & --2.50 \\
triviaqa               & 90.98 
& \textbf{91.71} & 90.89 & \underline{91.35} & 90.37 & --0.52 & 89.72 & \textbf{89.61} & \underline{88.60} & 88.52 & 88.47 & --0.13 \\
lsht                   & 46.50 & \textbf{46.00} & 39.50 & 42.00 & \underline{44.00} & +4.50 & 40.75 & \textbf{44.25} & 35.00 & 38.20 & \underline{40.53} & +5.53 \\
passage\_count         &  9.50 & \textbf{ 8.72} & \underline{ 4.59} &  4.39 &  2.50 & --2.09 & 10.00 & \textbf{ 4.63} & \underline{ 4.00} & 
\underline{ 4.00} & \underline{ 4.00} & 0.00 \\
passage\_retrieval\_en & 99.50 & \underline{94.19} & 82.00 & 83.00 & \textbf{94.50} & +12.50 &100.00 & \textbf{98.17} & 75.00 & 85.50 & \underline{88.50} & +13.50 \\
passage\_retrieval\_zh & 97.29 & 79.86 & \underline{90.67} & 90.42 & \textbf{93.83} & +3.16 & 96.50 & 86.18 & 85.00 & 86.00 & \textbf{89.50} & +4.50 \\
lcc                    & 53.69 & 54.24 & \underline{54.30} & 54.15 & \textbf{55.09} & +0.79 & 60.22 & \underline{59.70} & 59.59 & 58.64 & \textbf{60.08} & +0.49 
\\
repobench-p            & 51.28 & 52.35 & \textbf{54.10} & \underline{53.90} & 53.60 & --0.50 & 65.78 & \textbf{64.72} & 60.56 & 61.55 & \underline{63.33} & +2.77 \\
\midrule
\textbf{Mean} & 48.82 & 41.40 & 46.92 & \underline{47.12} & \textbf{47.77} & \textbf{+0.85}
              & 48.80 & 38.95 & 45.57 & \underline{46.45} & \textbf{47.23} & \textbf{+1.66} \\
\bottomrule
\end{tabular}%
}
\caption{Full LongBench results (4K--16K context).
FA = FlashAttention; MInf = MInference. 
CRISP retains the highest mean score, while FP 0.97 achieves only marginal gains over 
FP 0.95 at the cost of higher latency.
$\Delta$ is vs FP 0.95.}
\label{tab:longbench}
\end{table*}

\begin{table*}[h]
\centering
\small
\begin{tabular}{lccccccc}
\toprule
\textbf{Model / Method} & \textbf{4k} & \textbf{8k} & \textbf{16k} &
    \textbf{32k} & \textbf{64k} & \textbf{131k} & \textbf{Avg} \\
\midrule
\multicolumn{8}{l}{\textit{Llama-3.1-8B-Instruct}} \\
FlashAttn (FA)     & 95.67 & 93.99 & 93.75 & 86.78 & 84.62 & 79.33 & 89.02 \\
MInference         & \underline{95.67} & \textbf{94.23} & \textbf{94.47} & 86.30 & 83.65 & 62.50 & 86.14 \\
FP $\gamma{=}0.95$ & 94.95 & \textbf{94.23} & 93.99 & \textbf{90.63} & 83.41 & \textbf{78.12} & \textbf{89.22} \\
FP $\gamma{=}0.97$ & 94.95 & 93.75 & \underline{94.23} & \underline{89.91} & \underline{84.38} & \underline{77.64} & \underline{89.14} 
\\
CRISP (ours)       & \textbf{95.91} & \underline{93.99} & 93.03 & 87.98 & \textbf{86.06} & 75.96 & 88.82 \\
\midrule
\multicolumn{8}{l}{\textit{Qwen2.5-7B-Instruct}} \\
FlashAttn (FA)     & 94.47 & 93.03 & 89.18 & 87.02 & 66.11 & 25.24 & 75.84 \\
MInference         & \textbf{94.47} & \textbf{93.51} & 89.18 & 86.54 & 53.12 & 17.79 & 72.44 \\
FP $\gamma{=}0.95$ & \underline{94.23} & 91.59 & \underline{89.66} & 84.62 & \underline{66.34} & \textbf{26.68} & 75.52 \\
FP $\gamma{=}0.97$ & 93.99 & 92.31 & 88.94 & \underline{87.02} & 66.11 & \underline{26.44} & \underline{75.80} \\
CRISP (ours) 
      & 93.99 & \underline{92.55} & \textbf{90.63} & \textbf{87.26} & \textbf{67.79} & 25.00 & \textbf{76.20} \\
\bottomrule
\end{tabular}
\caption{Full RULER results per context length.
CRISP wins Qwen
outright over all methods and maintains competitive performance on Llama.}
\label{tab:ruler}
\end{table*}

\end{document}